%% file: xiaomi.tex
\documentclass[11pt, a4paper, logo, copyright, nonumbering]{xiaomi}

\usepackage[authoryear, sort&compress, round]{natbib}
\usepackage{dblfloatfix}
\usepackage{ulem}
\usepackage{caption}
\usepackage{dramatist}
\usepackage{xspace}
\usepackage{pifont} 
\usepackage{multirow}
\usepackage{tcolorbox}
\usepackage{xltabular}
\usepackage{longtable}
\usepackage{hyperref}
\usepackage{wrapfig}
\usepackage{graphicx}

\usepackage{amsfonts}
\usepackage{amsmath}
\usepackage{amssymb}
\usepackage{lineno}
\usepackage{multirow}
\usepackage{adjustbox}

\usepackage[bottom]{footmisc}

\usepackage{CJKutf8}
\usepackage{subfigure}
\usepackage{setspace}
\usepackage{makecell}
\usepackage{graphicx}
\usepackage{subcaption}
\usepackage{multicol} 

\newcommand{\mimo}{MiMo-7B}
\newcommand{\mimobase}{MiMo-7B-Base}
\newcommand{\mimozero}{MiMo-7B-RL-Zero}
\newcommand{\mimorl}{MiMo-7B-RL}
\newcommand{\rlsys}{Seamless Rollout Engine}
\newcommand{\rone}{Deepseek-R1}
\newcommand{\codereward}{test difficulty driven reward}

\definecolor{xiaomiorange}{HTML}{FF6901}

\title{\centering MiMo: Unlocking the Reasoning Potential of Language Model – From Pretraining to Posttraining}

\author{
 LLM-Core Xiaomi
}

\begin{abstract}

We present \mimo{}, a large language model born for reasoning tasks, with optimization across both pre-training and post-training stages.
During pre-training, we enhance the data preprocessing pipeline and employ a three-stage data mixing strategy to strengthen the base model's reasoning potential. 
\mimobase{} is pre-trained on 25 trillion tokens, with additional Multi-Token Prediction objective for enhanced performance and accelerated inference speed.
During post-training, we curate a dataset of 130K verifiable mathematics and programming problems for reinforcement learning, integrating a test-difficulty–driven code-reward scheme to alleviate sparse-reward issues and employing strategic data resampling to stabilize training.
Extensive evaluations show that \mimobase{} possesses exceptional reasoning potential, outperforming even much larger 32B models.
The final RL-tuned model, \mimorl{}, achieves superior performance on mathematics, code and general reasoning tasks, surpassing the performance of OpenAI o1-mini.
The model checkpoints are available at \textcolor{xiaomiorange} 
{\url{https://github.com/xiaomimimo/MiMo}}.

\end{abstract}

\begin{document}
\maketitle

\vspace{0.8cm}

\begin{figure}[h]
    \centering
    \includegraphics[width=0.99\textwidth]{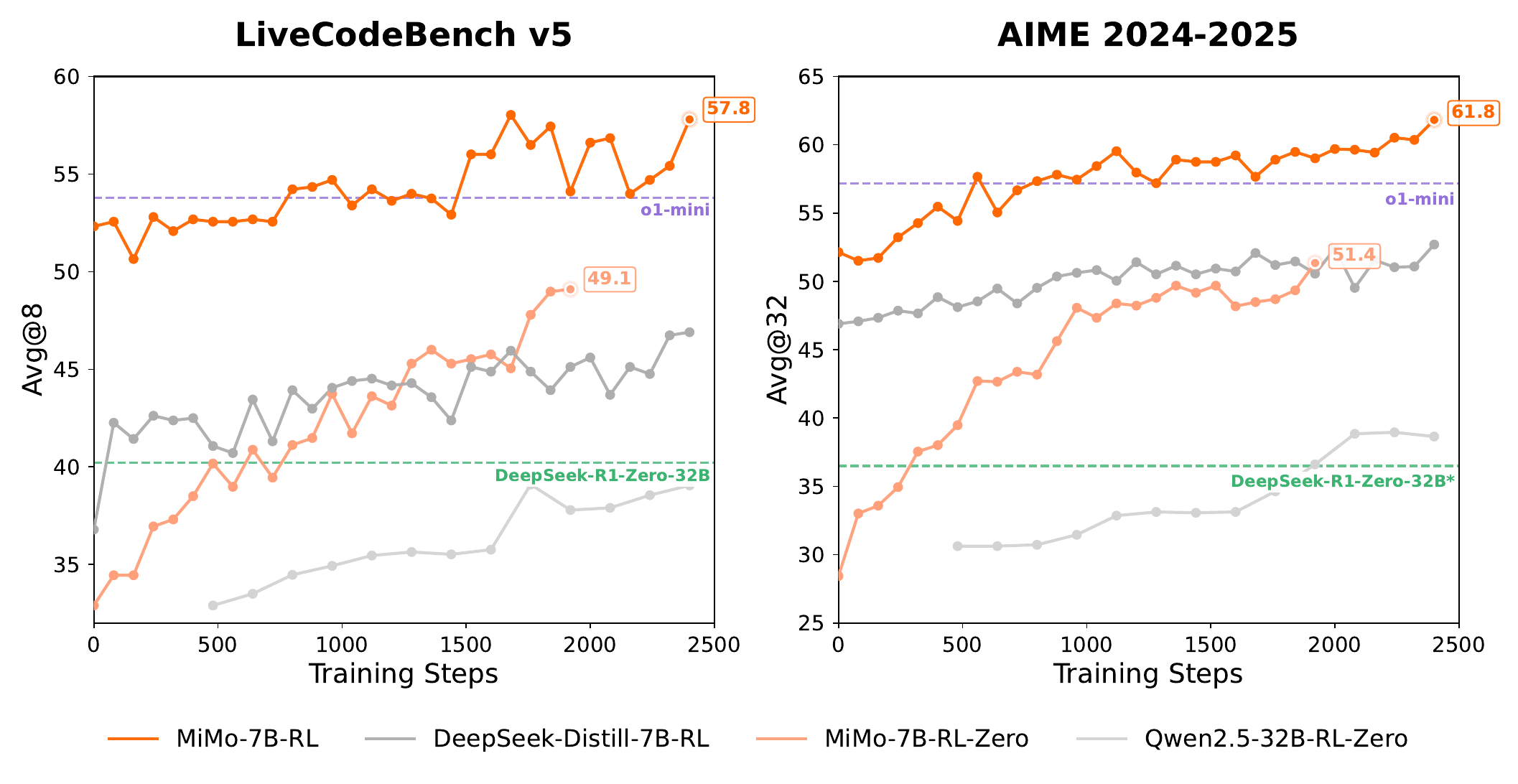}
    \caption{
    Performance of \mimo{} in code and math reasoning benchmark.
    }
    \label{fig:teaser}
\end{figure}

\newpage

\begin{spacing}{0.9}
\tableofcontents
\end{spacing}

\newpage

\section{Introduction}

Large language models (LLMs) with advanced reasoning capabilities, such as OpenAI o-series~\citep{o1}, DeepSeek R1~\citep{guo2025deepseek}, and Claude 3.7~\citep{claude3.7}, have achieved remarkable performance in complex tasks like mathematical reasoning and code generation.
Through large-scale reinforcement learning (RL), these models develop sophisticated reasoning patterns, including step-by-step analysis, self-reflection and backtracking, enabling more robust and accurate problem solving capabilities across diverse domains.
This emerging paradigm represents a significant advancement in artificial intelligence's approach for tackling intricate challenges.

Currently, most successful RL works, including open-source research, rely on relatively large base models, e.g., 32B models, particularly for enhancing code reasoning capabilities.
Moreover, it was widely considered that achieving uniform and simultaneous improvements in both mathematical and code capabilities within a small model is challenging.
Nonetheless, we believe that the effectiveness of the RL trained reasoning model relies on the inherent reasoning potential of the base model.
To fully unlock the reasoning potential of language models, efforts must focus not only on post-training but also on pre-training strategies tailored to reasoning.

In this work, we present \mimo{}, a series of models trained from scratch and born for reasoning tasks.
Our RL experiments from \mimobase{} show that our model possesses extraordinary reasoning potential, even outperforming much larger 32B models.
Additionally, we perform RL training on a cold-started SFT model, resulting in \mimorl{}, which demonstrates superior performance on both mathematics and code reasoning tasks, surpassing the performance of OpenAI o1-mini.
Here are our detailed contributions:

\paragraph{Pre-Training: Base Model Born for Reasoning}
\begin{itemize}
    \item We optimize data preprocessing pipeline, enhancing text extraction toolkits and applying multi-dimensional data filtering to increase reasoning pattern density in pre-training data. We also employ multiple strategies to generate massive diverse synthetic reasoning data.
    \item We adopt a three-stage data mixture strategy for pre-training. Overall, \mimobase{} is pre-trained on approximately 25 trillion tokens.
    \item We incorporate Multiple-Token Prediction as an additional training objective, which enhances model performance and accelerates inference.
\end{itemize}

\paragraph{Post-Training Recipe: Pioneering Reasoning Model}
\begin{itemize}
    \item We curate 130K mathematics and code problems as RL training data, which can be verified by rule-based verifiers. Each problem undergoes careful cleaning and difficulty assessment to ensure quality. We employ only rule-based accuracy rewards to avoid potential reward hacking.
    \item To mitigate the sparse reward issue for challenging code problems, we introduce a test difficulty driven code reward. By assigning fine-grained scores for test cases with varying difficulty levels, the policy can be more effectively optimized via dense reward signal.
    \item We implement a data re-sampling strategy to enhance rollout sampling efficiency and stabilize policy updates, particularly in the later phases of RL training.
\end{itemize}

\paragraph{RL Infrastructures}
\begin{itemize}
    \item We develop a \rlsys{} to accelerate RL training and validation. Our design integrates continuous rollout, asynchronous reward computation, and early termination to minimize GPU idle time, achieving 2.29$\times$ faster training and 1.96$\times$ faster validation.
    \item We support MTP in vLLM and enhance the robustness of the inference engine in RL system.
\end{itemize}

\paragraph{Summary of Evaluation Results}
\begin{itemize}
    \item \textbf{\mimobase{}} outperforms SoTA open-source models of approximately 7B parameters, excelling in general knowledge and coding tasks. On BBH, it achieves a score of 75.2, showcasing superior reasoning capabilities. Its strong performance on SuperGPQA further highlights its ability to handle complex graduate-level questions.
    \item \textbf{\mimozero{}} surpasses the RL training performance of the 32B base model on both mathematics and code tasks. This underscore its efficiency and potential in RL training, positioning \mimo{} as a compelling candidate for future advancements in RL.
    \item \textbf{\mimorl{}} achieves excellent reasoning performance. It scores 55.4 on AIME 2025, exceeding o1-mini by 4.7 points. In algorithm code generation tasks, \mimorl{} demonstrates extremely impressive results, significantly outperforming OpenAI o1-mini on both LiveCodeBench v5 and the latest v6, demonstrating robust and stable capabilities. \mimorl{} also maintains competitive general performance.
\end{itemize}

\paragraph{Open-Source}
We open-source \mimo{} series, including checkpoints of the base model, SFT model, RL model trained from base model, and RL model trained from the SFT model.
We believe this report along with the models will provides valuable insights to develop powerful reasoning LLM that benefit the larger community.

\section{Pre-Training}
\label{sec:pretrain}

In this section, we first detail our strategies to enhance reasoning capabilities during \mimo{} pre-training process, encompassing pre-training data construction, model architecture design, and hyper-parameter settings. 
Then we demonstrate the reasoning potential of \mimobase{} model.

\subsection{Pre-Training Data}
\label{sec:data}

The pre-training corpus for \mimo{} integrates diverse sources, including web pages, academic papers, books, programming code, and synthetic data. 
We believe that incorporating more data with high-quality reasoning patterns during pre-training stage can substantially enhance the reasoning potential of the resulting language model.
To achieve this goal, we first optimize our natural text preprocessing pipeline to improve quality and most importantly, reasoning data density.
Second, we leverage advanced reasoning models to generate extensive synthetic reasoning data.
Finally, we implement a three-stage data mixture strategy to maximize our model's reasoning potential across various tasks and domains.

\paragraph{Better Reasoning Data Extraction} 
Web pages naturally contain content with high density reasoning patterns, such as coding tutorial and mathematics blogs.
However, we discover that commonly used extractors~\citep{barbaresi-2021-trafilatura} often fail to preserve mathematics equations and code snippets embedded in the webpage.
To address this limitation, we develop a novel HTML-extraction tool specially optimized for mathematics content~\citep{liu2024finemath,paster2023openwebmath,zhou2025megamath}, code blocks, and forum websites.
For papers and books, we enhance PDF parsing toolkits to better handle STEM and code content.
With these optimized extraction tools, we successfully preserved massive reasoning patterns for subsequent processing stages.

\paragraph{Fast Global Deduplication}
Data deduplication plays an important role in improving training efficiency and reducing overfitting. 
We adopt both URL deduplication and MinHash deduplication~\citep{broder1997resemblance} across all webpage dumps.
Through extreme engineering optimization, we can complete this global deduplication process within a single day.
Since deduplication algorithms treat high-quality and low-quality text equally without content awareness, we subsequently adjust the final data distribution according to multi-dimension quality scores.

\paragraph{Multi-Dimensional Data Filtering}
High-quality pre-training data with rich reasoning patterns is crucial for developing models with strong reasoning capabilities.
We find that commonly used heuristic rule-based filters~\citep{penedo2023refinedweb,penedo2024fineweb} incorrectly filter high-quality web pages containing substantial mathematical and code content.
To address this limitation, we instead fine-tune small LLMs to serve as data quality taggers, performing domain classification and multi-dimensional quality assessment.

\paragraph{Synthetic Reasoning Data}
Another crucial source for reasoning patterns is synthetic data generated by advanced reasoning models.
We employ multiple strategies to generate diverse synthetic reasoning responses.
First, we select STEM content tagged with high reasoning depth and prompt models to develop insightful analyses and perform in-depth thinking based on the source materials.
Second, we gather mathematics and code problems and prompt reasoning models to solve them.
Additionally, we incorporate general domain queries, particularly creative writing tasks.
Notably, our preliminary experiments reveal that, unlike non-reasoning data, synthetic reasoning data can be trained for extremely high number of epochs without overfitting risk.

\paragraph{Three-Stage Data Mixture} 
To optimize the pre-training data distribution, we adopt a three-stage data mixture strategy in the final model training:
\begin{itemize}
    \item \textbf{Stage 1}: We incorporate all data sources except synthetic responses for reasoning task queries. We downsample overrepresented content, such as ads, news, job postings, and materials with insufficient knowledge density and reasoning depth. We also upsample high-value data from professional domains with superior quality.
    \item \textbf{Stage 2}: Building on the curated distribution in Stage 1, we significantly increase mathematics and code related data to $\sim$70\% of the mixture. This approach is expected to enhance specialized skills without compromising general language abilities~\citep{zhu2024deepseek}. The first two stages are trained with an 8,192-token context length.
    \item \textbf{Stage 3}: To boost the capabilities for solving complex tasks, we further incorporate $\sim$10\% synthetic responses for mathematics, code, and creative writing queries. Simultaneously, we extend the context length from 8,192 to 32,768 in the final stage.
\end{itemize}

Through this process, we build a large high-quality pre-training dataset comprising approximately \textbf{25 trillion} tokens.

\subsection{Model Architecture}
\label{sec:pretrain:modelarc}

\begin{figure}[t]
    \centering
    \includegraphics[width=0.99\textwidth]{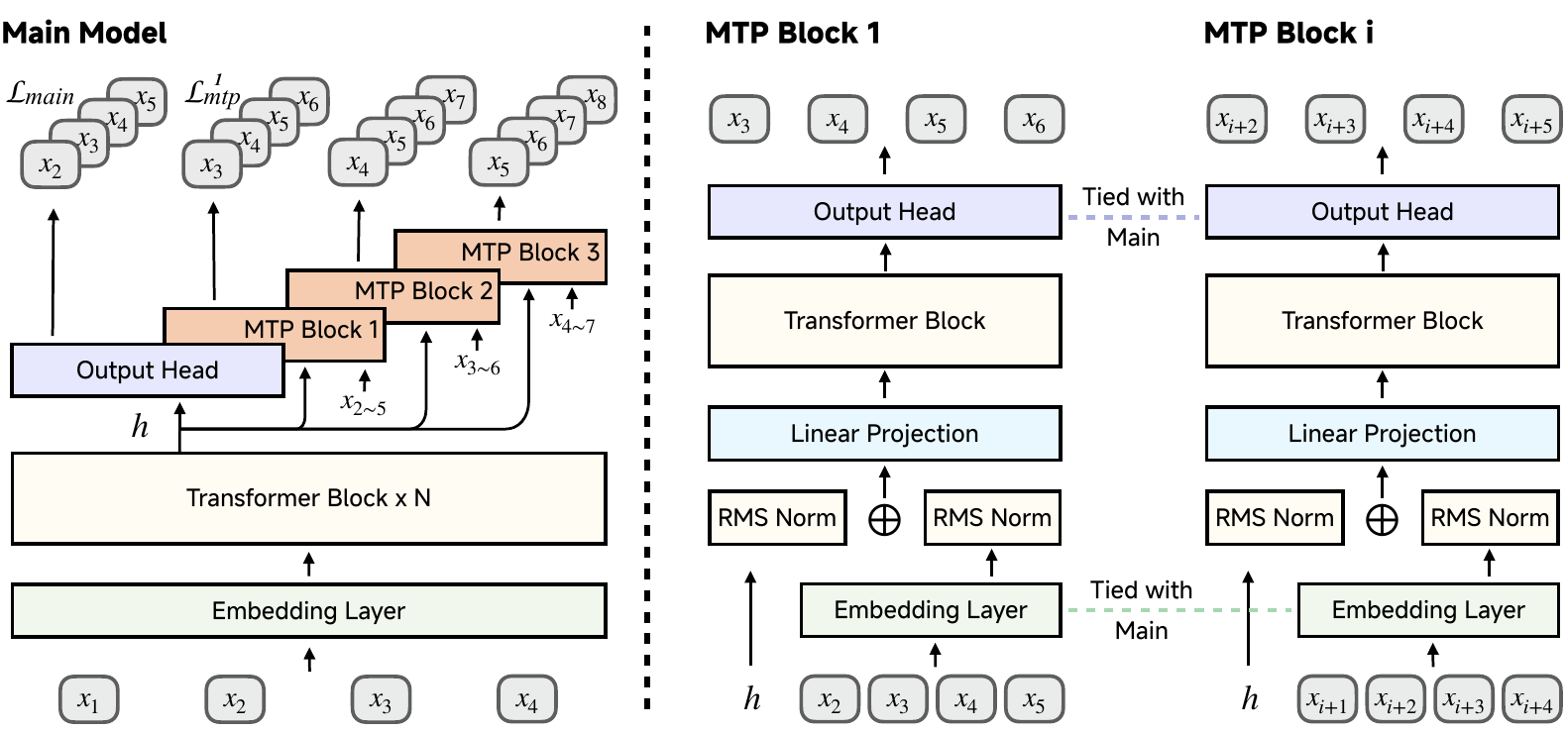}
    \caption{Implementation of Multi-Token Prediction with \mimo{}. During pre-training we use a single MTP layer, while the inference stage can use multiple MTP layers for additional speedup.}
    \label{fig:model_arch}
\end{figure}

\mimo{} follows the general decoder-only Transformer architecture~\citep{vaswani2017attention,radford2018improving}, and consists of Grouped-Query Attention (GQA, ~\citealt{ainslie2023gqa}), pre-RMSNorm~\citep{zhang2019root}, SwiGLU activation~\citep{dauphin2017language} and Rotary Positional Embedding (RoPE, ~\citealt{su2024roformer}), similar to Llama~\citep{touvron2023llama,grattafiori2024llama} and Qwen~\citep{yang2024qwen2}.

Reasoning models often face an inference speed bottleneck due to their lengthy auto-regressive generation process, despite the high correlation and predictability observed among consecutive tokens in their reasoning paths.

\paragraph{MTP Modules}
Inspired by DeepSeek-V3~\citep{liu2024deepseek}, we incorporate Multi-Token Prediction (MTP)~\citep{gloeckle2024better} as an additional training objective.
This approach enables the model to strategically pre-plan and generate representations that facilitate more accurate and potentially faster prediction of future tokens.
As shown in Figure~\ref{fig:model_arch}, we implement distinct MTP setups for pre-training and inference.
During pre-training, we utilize only a single MTP layer, as our preliminary studies show that multiple MTP layers yield no further improvement.
In contrast, we find that multiple parallel MTP layers significantly accelerate inference through speculative decoding.
To implement this, after pre-training, we replicate the pre-trained single MTP layer into two identical copies.
Then, with the main model and first MTP layer frozen, we fine-tune two new MTP layers for inference speedup.

\paragraph{MTP Inference Speedup} 
During inference, these MTP layers can be utilized for speculative decoding~\citep{leviathan2023fast,xia2022speculative} to reduce generation latency. 
We evaluated the performance of the MTP layers on the AIME24 benchmark. 
The first MTP layer achieves a remarkably high acceptance rate about 90\%, while even the third MTP layer maintains an acceptance rate above 75\%.
This high acceptance rate enables \mimo{} to deliver enhanced decoding speed, particularly in reasoning scenarios requiring extremely long outputs.

\subsection{Hyper-Parameters}

\paragraph{Model Hyper-Parameters} 
We set the number of Transformer layers to 36 and the hidden dimension to 4,096. The intermediate hidden dimension of FFN is set to 11,008. The number of attention heads is 32 and there are 8 key-value groups.

\paragraph{Training Hyper-Parameters} 
For optimization, we use AdamW~\citep{loshchilov2017decoupled} with $\beta_1 = 0.9$, $\beta_2 = 0.95$, and weight decay of 0.1. We apply gradient clipping with a maximum norm of 1.0.

During the first two pre-training stages, the maximum sequence length is 8,192 tokens with the RoPE base of 10,000. 
Stage 3 expands these parameters to 32,768 tokens and 640,000, respectively.

Our learning rate schedule begins in Stage 1 with a linear warmup from 0 to $1.07\times10^{-4}$ over the first 84B tokens, followed by a constant phase at $1.07\times10^{-4}$ for 10.2T tokens, and concludes with a cosine decay to $3\times10^{-5}$ over 7.5T tokens. 
This rate of $3\times10^{-5}$ is maintained throughout Stage 2 (4T tokens) and for the first 1.5T tokens of Stage 3. Subsequently, the learning rate decays via a cosine schedule to $1\times10^{-5}$ over the final 500B tokens.

We implement a linear batch size warmup to 2,560 over the first 168B tokens and maintain this value throughout the remainder of Stage 1 and Stage 2.
In Stage 3, the batch size is fixed at 640.

The MTP loss weight is set to 0.3 for the first 10.3T tokens, then reduced to 0.1 for the remainder of pre-training.

\subsection{Pre-Training Evaluation}

\subsubsection{Evaluation Setup}

We evaluate \mimobase{} on a series of benchmarks, encompassing natural language understanding and reasoning, scientific question answering, reading comprehension, mathematics reasoning, coding, Chinese understanding, and long-context comprehension capabilities:

\textbf{Language understanding and reasoning}: BBH~\citep{suzgun2022challenging}, MMLU~\cite{hendrycks2020measuring}, MMLU-Redux~\citep{gema2024we}, MMLU-Pro~\citep{wang2024mmlu}, ARC~\citep{clark2018think}, HellaSwag~\citep{zellers2019hellaswag}, PIQA~\citep{bisk2020piqa}.

\textbf{Closed-book question answering}: TriviaQA~\citep{joshi2017triviaqa}, NaturalQuestions~\citep{kwiatkowski2019natural}.

\textbf{Scientific question answering}: GPQA~\citep{rein2024gpqa}, SuperGPQA~\citep{du2025supergpqa}.

\textbf{Reading comprehension}: DROP~\citep{dua2019drop}, RACE~\citep{lai2017race}.

\textbf{Mathematics reasoning}: AIME~\citep{AIME}, GSM8K~\citep{cobbe2021training}, MATH~\citep{hendrycks2021measuring}.

\textbf{Coding}: LiveCodeBench~\citep{jain2024livecodebench}, HumanEval~\citep{chen2021evaluating}, HumanEval+~\citep{liu2023your}, MBPP~\citep{austin2021program}, MBPP+~\citep{liu2023your}, CRUXEval~\citep{gu2024cruxeval}.

\textbf{Miscellaneous}: WinoGrande~\citep{sakaguchi2021winogrande}, AGIEval~\citep{zhong2023agieval}.

\textbf{Chinese understanding}: C-Eval~\citep{huang2023c}, CMMLU~\citep{li2023cmmlu}.

\textbf{Long-Context Comprehension}: RULER~\citep{hsieh2024ruler}

We compare \mimobase{} with other open-source base models of comparable size, including Llama-3.1-8B~\citep{grattafiori2024llama}, Gemma-2-9B~\citep{gemmateam2024gemma2}, and Qwen2.5-7B~\citep{yang2024qwen2}. 
The evaluation of all models shares the same evaluation settings.

\subsubsection{Upper Bounds of Reasoning Capability}

Traditional evaluation methods often underestimate a model's true reasoning potential by relying on single-pass success rates or average performance across multiple samplings.
Following ~\citet{yue2025doesreinforcementlearningreally}, we adopt the pass@k metric, which considers a problem solved if any of k sampled solution is correct, to better assess the reasoning capacity boundary of different models.

\begin{figure}[t]
    \centering
    \includegraphics[width=0.99\textwidth]{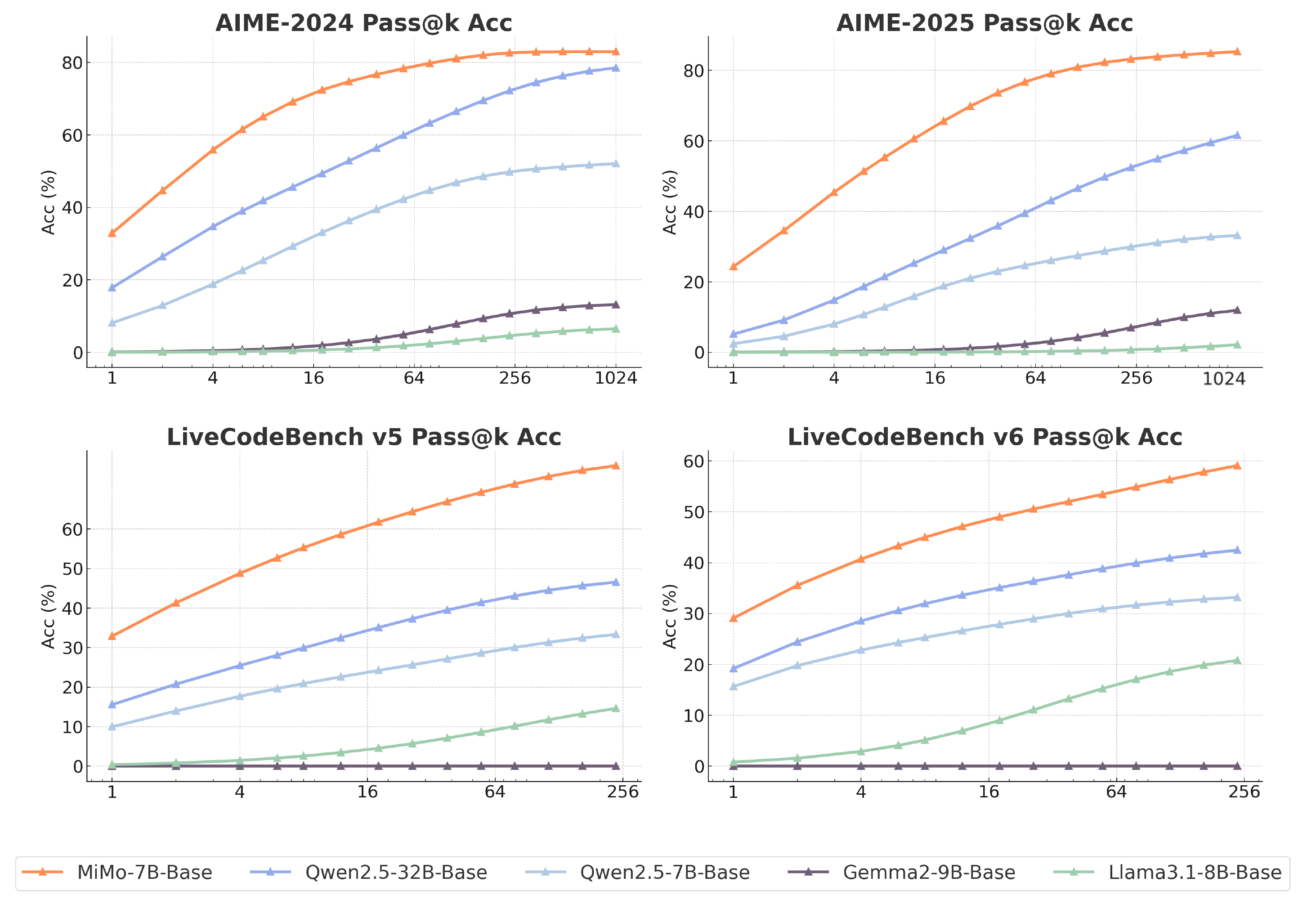}
    \caption{Pass@k curves of different base models across multiple reasoning benchmarks.}
    \label{fig:passk}
\end{figure}

As illustrated in Figure~\ref{fig:passk}, \mimobase{} achieves significantly higher pass@k scores than all compared models, including the 32B baseline, across all benchmarks and evaluated k values.
Notably, the performance gap between \mimobase{} and other baselines widens steadily as k increases, particularly on LiveCodeBench.
These results demonstrates the superior reasoning potential of \mimobase{}, which establishes a strong base policy for RL training.

\subsubsection{Evaluation Results}

\input{tables/base_eval}

\paragraph{General Reasoning} 
\mimobase{} achieves superior performance in general knowledge and reasoning, outperforming open-source models of comparable size. 
On BBH, a benchmark evaluating language reasoning abilities, \mimobase{} scores 75.2, surpassing Qwen2.5-7B by about 5 points.
Furthermore, SuperGPQA results show our model's robust performance in solving graduate-level problems.
On DROP, a reading comprehension benchmark, \mimobase{} outperforms compared models, showing advanced language understanding capability.

\paragraph{Code and Mathematics Reasoning} 
\mimobase{} demonstrates strong proficiency in coding and mathematics tasks. 
On LiveCodeBench v5, it scores 32.9, far surpassing Llama-3.1-8B and Qwen-2.5-7B. 
Similarly, on AIME 2024, our model achieves 32.9, significantly outperforming other comparably sized base models. 
These results highlight \mimobase{}’s extraordinary problem-solving abilities and its huge potential for complex reasoning tasks.

\begin{figure}[t]
    \centering
    \includegraphics[width=0.99\textwidth]{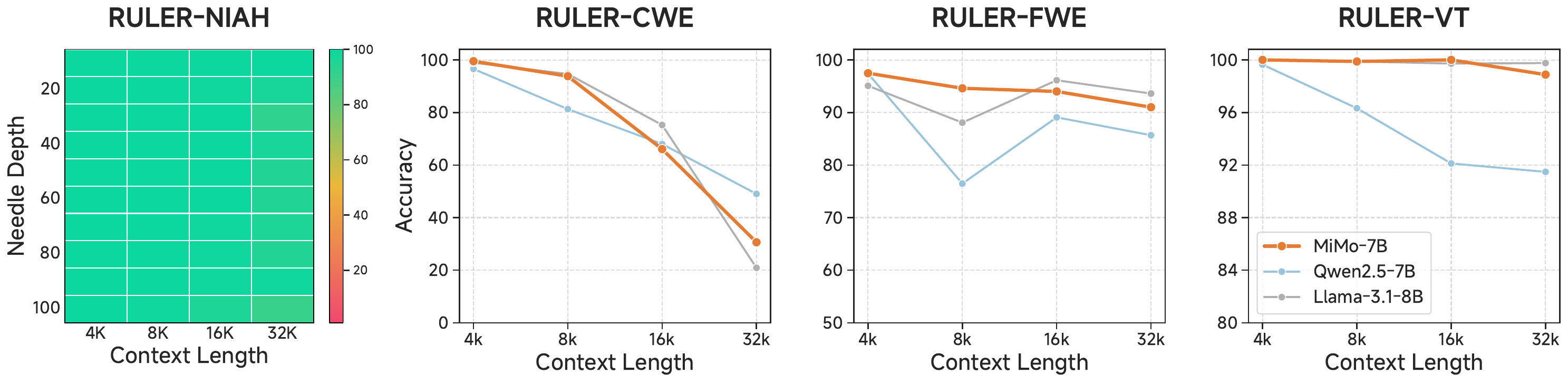}
    \caption{Results of long-context comprehension on RULER. Our \mimobase{} achieves near-perfect NIAH retrieval performance within the supported 32K context length, and delivers remarkable performance on Common Words Extraction (CWE), Frequent Words Extraction (FWE), and Variable Tracking (VT) that emphasizes long-context reasoning beyond retrieval.}
    \label{fig:lctx_ruler}
\end{figure}

\paragraph{Long-Context Comprehension} 
The ability to understand and reason over long contexts is essential for modern thinking models~\citep{liu2025comprehensive}, as it enables them to produce long and complex reasoning chains. 

For the needle-in-a-haystack (NIAH) tasks (Single, Multi-keys, Multi-values, and Multi-queries NIAH) that focus on long-context retrieval, we aggregate their accuracy across varying depths and context lengths, as depicted in the leftmost panel of Figure~\ref{fig:lctx_ruler}. We observe that \mimo{} achieves near-perfect retrieval performance across all positions within the 32K context window.

Beyond pure retrieval, \mimo{} excels in tasks requiring long-context reasoning, including Common Words Extraction (CWE), Frequent Words Extraction (FWE), and Variable Tracking (VT). It delivers remarkable performance and surpasses Qwen2.5-7B in most scenarios. These results validate the efficacy of our strategy to incorporate diverse data with high-quality reasoning patterns during pre-training.

\section{Post-Training}
\label{sec:pt}

After the pre-training stage, post-training are implemented on \mimobase{}.
Specifically, we develop \mimozero{} through direct RL from \mimobase{}, and \mimorl{} trained from an SFT version of \mimo{}.

\subsection{Supervised Fine-Tuning}

\paragraph{SFT Data}

The SFT data consists of a combination of open-source and proprietary distilled data. To ensure optimal quality and diversity, we implement a three stage preprocessing pipeline.
First, we eliminate all training queries that have 16-gram overlap with evaluation benchmarks to prevent data leakage. 
Then, we exclude samples with mixing language or incomplete response.
Finally, we capped the number of responses per query at eight, striking a balance between preserving diversity and preventing redundancy.
Following this preprocessing, our final SFT dataset comprises about 500K samples.

\paragraph{SFT Hyper-parameters}

We fine-tune the \mimobase{} model with a constant learning rate of $3\times10^{-5}$ and batch size of 128. Samples are packed to the maximum length of 32,768 tokens during training.

\subsection{RL Data Curation}

We utilize two categories of verifiable problems, mathematics and code, to formulate our RL training data.
Our preliminary studies demonstrate that high-quality problem sets plays a critical role in stabilizing the RL training process and further enhancing the LLM's reasoning capabilities.

\paragraph{Mathematical Data}
Our mathematical problem set is drawn from diverse sources, including open-source datasets and proprietary collected competition-level collections.
To mitigate the risk of reward hacking, we employ an LLM to filter proof-based and multiple-choice problems.
Unlike recent approaches that modify problems to ensure integer answers, we preserve original problems to minimize reward hacking.
Additionally, we perform global n-gram deduplication and carefully decontaminate of our problem set with evaluation benchmarks.

Model-based difficulty assessment is used to further improve the quality of our dataset.
Initially, we filter out problems that cannot be solved by advanced reasoning models, identifying those that are either too difficult or contain incorrect answers.
For the remaining problems, we rollout an SFT version of \mimo{} 16 times, eliminating problems with a passrate exceeding 90\%.
Notably, this process removes approximately 50\% of easy problems from the original problem set.
After data cleaning, we establish a mathematical training set comprising 100K problems.

\paragraph{Code Data}
For coding problems, we curate a high-quality training set comprising open-source datasets and our newly collected problem set.
We remove problems without test cases.
For problems with golden solutions, we exclude those where the golden solution failed to pass all test cases.
For problems without golden solution, we discard problems where no test case can be solved in 16 rollouts of advanced reasoning models.
Similar to math data, we utilize an SFT version of \mimo{} to filter out easy problems that are perfectly solved in all 16 rollouts.
This rigorous cleaning process yields 30K code problems.

During each RL iteration, we evaluate thousands of problems to compute the rewards, with each problem potentially containing hundreds of test cases. 
To improve reward computing efficiency and eliminate GPU idle time, we developed an online judge environment that enables parallel execution of extremely high-volume unit tests.

\paragraph{Reward Function}
We employ only rule-based accuracy rewards in our training process.
For mathematics data, we use the rule-based Math-Verify library to evaluate response correctness.
For code problems, we implement a \codereward{} as detailed in Section~\ref{sec:code_reward}.
No additional rewards, such as format reward and length penalty reward, is incorporated.

\subsection{RL Training Recipe}

We employ a modified version of Group Relative Policy Optimization (GRPO)~\citep{shao2024deepseekmath} with recently proposed improvement from the research community~\citep{hu2025open,yu2025dapo}.
For each problem $q$, the algorithm samples a group of responses $\left\{o_1,o_2,...,o_G\right\}$ from the old policy $\pi_{\theta_{old}}$, and update the policy $\pi_\theta$ by maximizing the following objective:
\begin{equation}
\begin{aligned}
\label{eq:grpo}
    \mathcal{J}_{\mathrm{GRPO}}\left(\theta\right)=& \mathbb{E}_{q\sim D,\{o_i\}_{i=1}^G\sim \pi_\theta(\cdot|q)} \\
    & \left[\frac{1}{\sum_{i=1}^G\left|o_i\right|}\sum_{i=1}^G \sum_{j=1}^{\left|o_i\right|} \mathrm{min}\left(\frac{\pi_\theta(o_i|q)}{\pi_{\theta_{old}}(o_i|q)}A_{i,j},\mathrm{clip}\left(\frac{\pi_\theta(o_i|q)}{\pi_{\theta_{old}}(o_i|q)},1-\varepsilon_\mathrm{low},1+\varepsilon_\mathrm{high}\right)A_{i,j}\right)\right]
\end{aligned}
\end{equation}
where $\varepsilon_{\mathrm{low}}$ and $\varepsilon_{\mathrm{high}}$ are hyper-parameters. $A_{i,j}$ is the advantage, which is computed by the rewards $\left\{r_1,r_2,...,r_G\right\}$ of responses in the same group:
\begin{equation}
    A_{i,j} = \frac{r_i-\mathrm{mean}(\{r_i\}_{i=1}^G)}{\mathrm{std}(\{r_i\}_{i=1}^G)}
\end{equation}

Upon the original GRPO algorithm, we incorporate several enhancements from recent research:
\begin{itemize}
    \item \textbf{Removal of KL Loss}~\citep{hu2025open,skywork-or1-2025}: simply removing the KL loss effectively unleashes the full potential of the policy model without compromising training stability.
    \item \textbf{Dynamic Sampling}~\citep{yu2025dapo}: in RL rollout phase, we over-sample and filter out prompts with passrate equal to 1 and 0, leaving all prompts in the batch with effective gradients while maintaining a consistent batch size. This strategy automatically calibrates problem difficulty throughout policy training.
    \item \textbf{Clip-Higher}~\citep{yu2025dapo}: we increase the upper clip bounds $\varepsilon_{\mathrm{high}}$ in Eq.~\ref{eq:grpo}, with a fixed lower clip bounds $\varepsilon_{\mathrm{low}}$. It can mitigate the entropy convergence problem and facilitate the policy to explore new solutions.
\end{itemize}

During training, we identify two key challenges affecting model performance: sparse rewards for code problems and diminishing sampling efficiency of dynamic sampling. 
Therefore, we propose \textbf{test complexity driven reward} function and \textbf{easy data re-sampling} approach, respectively.

\subsubsection{Test Difficulty Driven Reward}
\label{sec:code_reward}
Currently, for algorithm code generation tasks, existing RL works such as \rone{}~\cite{guo2025deepseek} adopt a rule-based reward strategy, where a solution is rewarded only if the generated code passes all the test cases for a given problem. However, for difficult algorithmic problems, the model might never receive any reward, preventing it from learning from these challenging cases and reducing training efficiency for dynamic sampling.

\paragraph{Various Test Difficulty in IOI Scoring Rules}
To address this limitation, we propose a new reward mechanism, \codereward{}. The design is inspired by the scoring rule of the International Olympiad in Informatics (IOI, ~\citealt{ioi}). In IOI contests, each complete problem is divided into multiple subtasks, and participants will obtain points for each subtask they successfully complete. Each subtask will have tests with different difficulty. 
Assigning different scores to subtasks better reflects how humans solve problems. 
For challenging problems, the model can still earn partial scores by solving some of the subtasks, which allows better utilization of these difficult examples during training.

\begin{figure}[t]
    \centering
    \includegraphics[width=\linewidth]{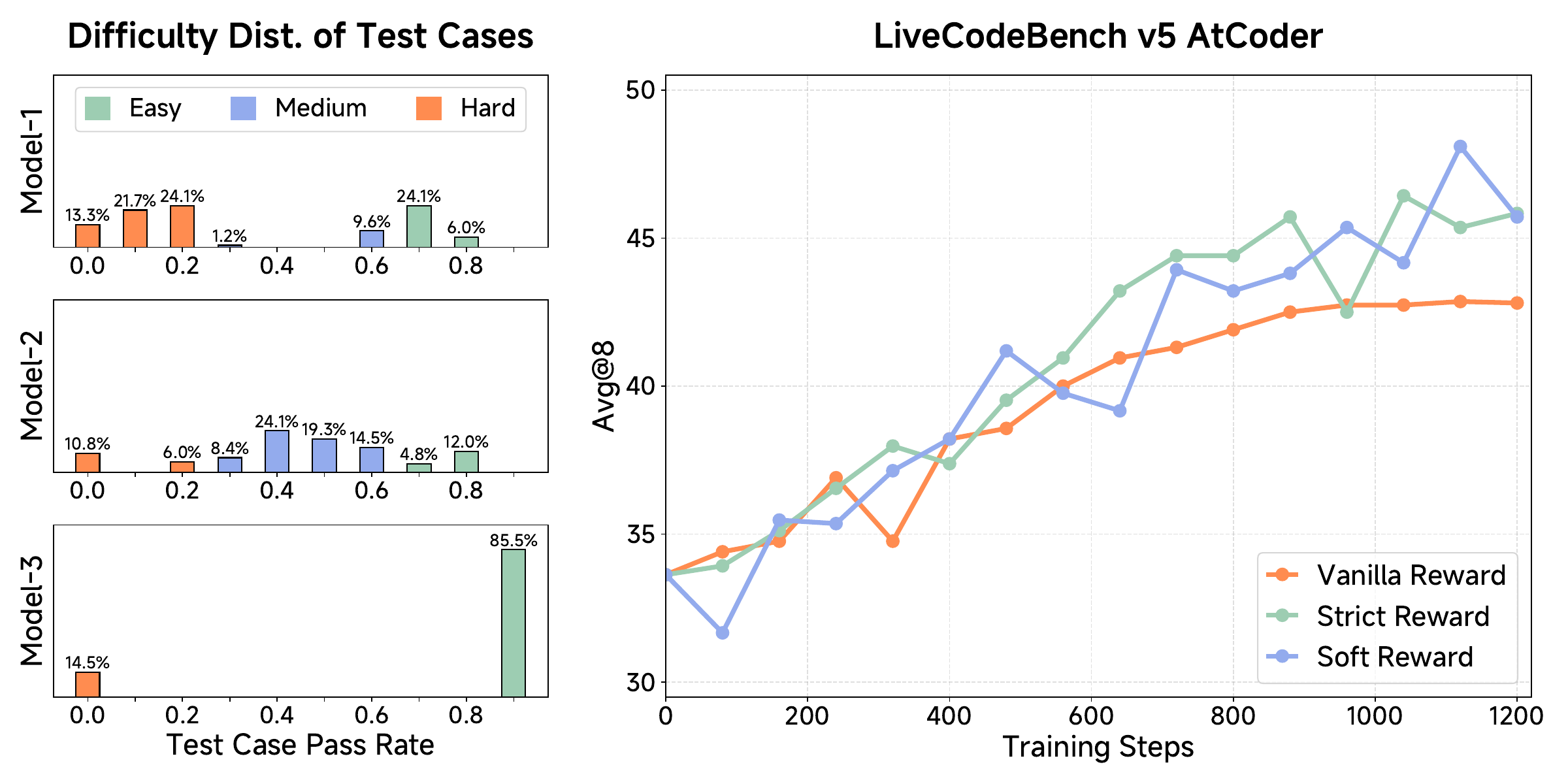}
    \caption{Experiments with test difficulty driven reward.}
    \label{fig:fine_test}
\end{figure}

\paragraph{Assigning Difficulty to Tests Based on Pass Rates}
We propose a technique for grouping test cases based on their difficulty. We utilize several models to perform multiple rollouts on each problem, and calculate the pass rate for each test case across all model-generated solutions. We then cluster the test cases into different difficulty levels according to their pass rates, with lower pass rates indicating higher difficulty.
The left part of Figure~\ref{fig:fine_test} presents the pass rates and difficulty levels for each test case of certain problem. The results reveal a clear stratification of test difficulty, and demonstrate that more capable models achieve higher pass rates.

\paragraph{Reward Rules}
After categorizing the tests into different difficulty levels, we design two reward schemes based on these difficulty levels: a strict scheme and a soft scheme. (1) \textit{Strict Reward.} Under the strict reward scheme, a solution receives the reward corresponding to a difficulty level only if it passes all tests in that group as well as in all lower-difficulty groups.
(2) \textit{Soft Reward.} In contrast, the soft reward scheme distributes the total score of each group equally among its tests. The final reward is the sum of the scores for all passed tests. The right part of Figure~\ref{fig:fine_test} compares the performance achieved by two reward schemes against the baseline without \codereward{}.

\subsubsection{Easy Data Filter and Re-Sampling}

During RL training, as the policy improves, an increasing number of problems achieve a perfect pass rate of 1.
Under dynamic sampling mechanism, these problems are subsequently filtered from the batch for policy update.
This filtration leads to drastic sampling efficiency degradation, as more rollouts are required to construct a batch of fixed size.
A straightforward approach to address this efficiency issue would be to entirely remove problems with perfect pass rates from the training data.
However, our preliminary studies show that this method introduces significant instability in policy updates.

To improve sampling efficiency without risking policy collapse, we developed an easy data re-sampling strategy.
During the training process, we maintain an easy data pool, where problems with perfect pass rates are stored.
When performing rollouts, there is a probability $\alpha$ (10\% in our experiments) to sample data from this easy data pool.
This strategy effectively stabilizes the policy update while improving sampling efficiency, especially in the later phases of RL training.

\subsubsection{Hyper-Parameters}
In our experiment, we employed a training batch size of 512, with an actor mini-batch size of 32. We executed 16 gradient updates per training iteration at a learning rate of 1e-6. The maximum sequence length was set to 32,768 tokens to facilitate complex reasoning tasks. During the training phase, both temperature and top-p parameters were configured at 1.0 to promote output diversity.

\input{sections/rlinfra}

\subsection{Post-Training Evaluation}

\subsubsection{Evaluation Setup}

We comprehensively evaluate reasoning models across a diverse range of benchmarks:

\textbf{Language understanding and reasoning}: MMLU-Pro~\citep{wang2024mmlu}.

\textbf{Scientific question answering}: GPQA Diamond~\citep{rein2024gpqa} with averaged score of 8 repetitions; SuperGPQA~\citep{du2025supergpqa}.

\textbf{Instruction following}: IFEval~\citep{zhou2023instructionfollowingevaluationlargelanguage} with averaged score of 8 repetitions.

\textbf{Reading comprehension}: DROP~\citep{dua2019drop}.

\textbf{Mathematics reasoning}: MATH500~\citep{lightman2023lets}; AIME 2024~\citep{AIME} and AIME 2025~\citep{AIME25} with averaged score of 32 repetitions.

\textbf{Coding}: LiveCodeBench v5 (20240801-20250201)~\citep{jain2024livecodebench} and LiveCodeBench v6 (20250201-20250501)~\citep{jain2024livecodebench} with averaged score of 8 repetitions.

During evaluation, we set the sampling temperature to 0.6 and top-p to 0.95 for all benchmarks. We set the maximum generation length to 32,768 tokens for mathematics reasoning, coding, and scientific question answering benchmarks, and to 8,192 tokens for other benchmarks.

We compare \mimorl{} against several strong baselines, including two non-reasoning models GPT-4o-0513, Claude-Sonnet-3.5-1022, and reasoning models OpenAI-o1-mini, QwQ-32B-Preview, DeepSeek-R1-Distill-Qwen-14B, and DeepSeek-R1-Distill-Qwen-7B.

\input{tables/rl_eval}

\subsubsection{Evaluation Results}

Table~\ref{tab:rl_eval} shows the evaluation results.
In mathematics reasoning, \mimorl{} achieves top-tier performance among models of comparable parameter sizes, trailing only slightly behind DeepSeek-R1-Distill-Qwen-14B on AIME 2024.
For algorithm code generation tasks, \mimorl{} demonstrates extremely impressive results.
On LiveCodeBench v5, it significantly outperforms OpenAI o1-mini, while on the latest LiveCodeBench v6, our model achieves a score of 49.3\%, surpassing QwQ-32B-Preview by over 10 points, demonstrating its robust and stable capabilities.
Notably, \mimorl{} also maintains strong general performance, exceeding both QwQ-32B-Preview and DeepSeek-R1-Distill-Qwen-7B, though we only include mathematics and code problems for RL.

We also presents the evaluation results for different version of \mimo{} in Table~\ref{tab:mimo_rl_result}.
\mimozero{} is trained from \mimobase{}, while \mimorl{} is trained from \mimo{}-SFT.
As shown, RL from the base model exhibits a stronger growth trend, improving from 32.9\% to on AIME 2024 for instance.
Nonetheless, RL training from the SFT model achieves a higher performance ceiling, attaining the best results across all evaluated benchmarks.

\begin{table}[h]
\centering
\small
\resizebox{0.88\textwidth}{!}{
\begin{tabular}{l|cc|cc}
\toprule
\textbf{Benchmark} & \textbf{MiMo-7B-Base} & \textbf{MiMo-7B-RL-Zero} & \textbf{MiMo-7B-SFT} & \textbf{MiMo-7B-RL} \\ 
\midrule
\multicolumn{5}{l}{\textbf{Mathematics}} \\
MATH500 & 37.4 & 93.6 & 93.0 & 95.8 \\ 
AIME 2024 & 32.9 & 56.4 & 58.7 & 68.2 \\ 
AIME 2025 & 24.3 & 46.3 & 44.3 & 55.4 \\ 
\midrule
\multicolumn{5}{l}{\textbf{Code}} \\
LiveCodeBench v5  & 32.9 & 49.1 & 52.3 & 57.8 \\ 
LiveCodeBench v6  & 29.1 & 42.9 & 45.5 & 49.3 \\ 
\bottomrule
\end{tabular}
}
   \caption{
        Evaluation results of MiMo-Series models on mathematics and coding benchmarks
    }
    \label{tab:mimo_rl_result}
\end{table}

\subsection{Discussion}

In this section, we share insights and observations from our exploration of \mimo{}'s post-training process, which we hope will benefit the research community.

\begin{wrapfigure}{r}{0.5\textwidth}
    \centering
    \includegraphics[width=0.99\linewidth]{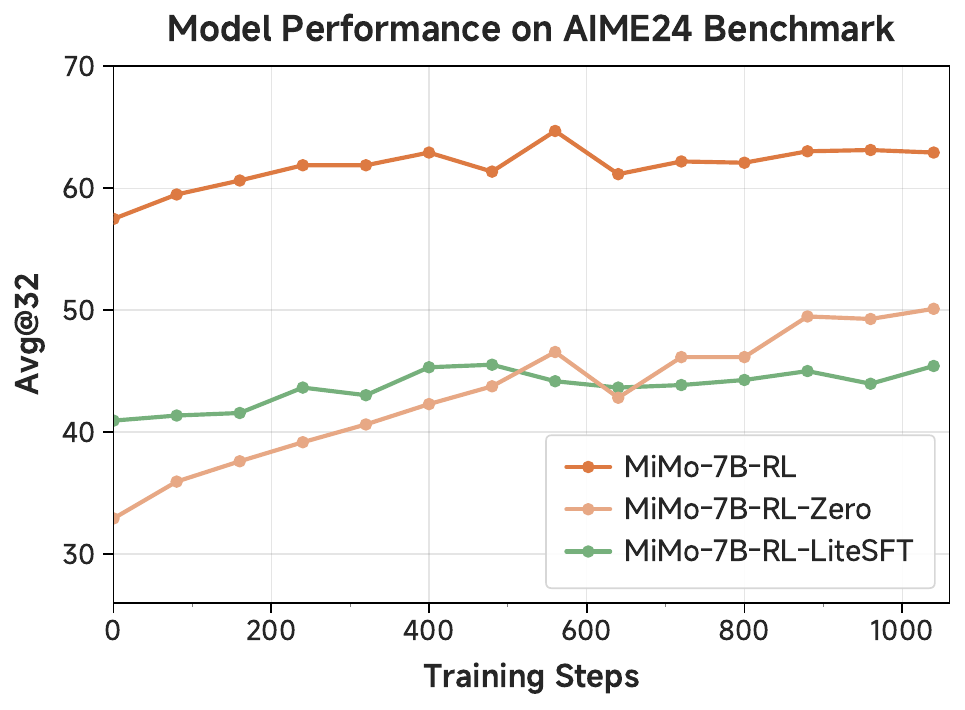}
    \caption{
    Performance comparison of three MiMo model variants during the RL process.
    }
    \label{fig:litesft}
\end{wrapfigure}

\paragraph{SFT for Format Alignment}
In the initial RL training steps from \mimobase{}, we observe that the model primarily learns to adapt the answer extraction function, e.g., ``\textbackslash boxed\{\}'' for mathematics problems.
Therefore, we investigate a ``light-weight'' SFT to help the base model align with the expected answer format.
However, as Figure~\ref{fig:litesft} demonstrates, the resulting \mimo{}-RL-LiteSFT model fails in both reasoning potential and final performance.
While \mimo{}-RL-LiteSFT begins with a higher performance than \mimozero{}, it falls behind the base model's trajectory after just 500 steps.
Furthermore, when compared to \mimorl{}, which undergoes ``heavier'' SFT, \mimo{}-RL-LiteSFT exhibits a similar growth trend but significantly underperforms due to its inferior starting point, ultimately leading to poorer final results.

\paragraph{Interference Between Different Domains}
During the later stages of RL training from \mimobase{}, maintaining a performance balance between mathematics and coding tasks proves challenging.
Between training steps 2000 and 2500, the model exhibits continuous improvement on code problems, while its performance on mathematical reasoning tasks fluctuates and declines.
In contrast, RL training on the cold-started SFT model shows consistent improvements across both domains.
Analysis of the model outputs reveals that the base model, with its strong exploration capabilities, tends to hack the reward for mathematics problems.
For code problems, however, the test-case-based verifier makes reward exploitation significantly harder.
This highlights the critical need for high-quality mathematical problem sets to ensure robust RL training.

\paragraph{Language Mixing Penalty}
Like DeepSeek-R1-Zero, we also observe language mixing issues during RL training on \mimobase{}.
To mitigate this problem, we introduce a language mixing penalty into the reward function.
However, we find designing such a penalty function is challenging.
While detecting Chinese characters in English responses is straightforward, the reverse is far more difficult, since mathematical equations and code inherently contain English words.
As a result, the penalty not only fails to fully resolve language mixing but also introduces the risk of reward hacking, such as always generating English responses regardless of the question language.

\paragraph{Impact of SFT Data Scaling}
Building upon preliminary experiments, our study significantly scaled the SFT dataset from approximately 500K to 6M instances. We empirically observed that this substantial expansion of SFT data resulted in marked improvements in the model's reasoning abilities and its capacity for generalized dialogue, without compromising its potential for subsequent RL. As detailed in Table~\ref{tab:mimo_7b_0530}, the model trained with 6M SFT instances exhibited considerable advancements over its counterpart trained with 500K instances in areas such as mathematical reasoning, code reasoning, scientific reasoning and general dialogue capabilities. Importantly, models subsequently fine-tuned with RL following this enhanced SFT stage also demonstrated sustained performance improvements.

\begin{table}[htbp!]
\centering
\resizebox{\columnwidth}{!}{%
\begin{tabular}{@{}lcccc@{}} 
\toprule
\textbf{Benchmark}         & \textbf{MiMo-7B-SFT-500K} & \textbf{MiMo-7B-SFT-6M} & \textbf{MiMo-7B-RL} & \textbf{MiMo-7B-RL-0530} \\
\midrule
AIME 24           & 58.7                      & 68.3                    & 68.2                & 80.1                     \\
AIME 25           & 44.3                      & 50.9                    & 55.4                & 70.2                     \\
MATH500             & 93.0                      & 94.8                    & 95.8                & 97.2                     \\
GPQA Diamond      & 50.7                      & 54.1                    & 54.4                & 60.6                     \\
LiveCodeBench v5   & 52.3                      & 53.4                    & 57.8                & 60.9                     \\
Alignbench v1.1   & 6.7                       & 7.1                     & 6.9                 & 7.4                      \\
\bottomrule
\end{tabular}%
}
\caption{Model Performance Comparison on Various Benchmarks.
MiMo-7B-RL-0530 was evaluated at a 48K context length, its training length, while the other three models were assessed at their 32K training context length. Evaluations for Alignbench v1.1~\cite{liu2024alignbenchbenchmarkingchinesealignment} were conducted using GPT-4.1 as the judge.}
\label{tab:mimo_7b_0530} 
\end{table}

\paragraph{On-Policy RL with Extended Generation Budget}
Our prior empirical investigations indicated that a vanilla implementation of GRPO was markedly prone to premature performance saturation. To mitigate this, we adopted an on-policy RL algorithm, drawing parallels with the approach used in MiMo-VL-7B-RL~\cite{coreteam2025mimovltechnicalreport}. Training with on-policy RL proved to be remarkably stable, while also enabling sustained growth in model efficacy throughout the learning process. Further extending our findings, we observed that continuous elevation of the generation length budget during on-policy RL training consistently boosts model performance. Specifically, our RL training protocol involved systematically increasing the model's generation length from 32K to 38K, and subsequently to 48K. This progressive extension of the generation budget was instrumental in our 7B model ultimately achieving parity with the Deepseek-R1 performance in mathematical reasoning. The MiMo-7B-RL-0530 model has been open-sourced and is publicly available\footnote{https://huggingface.co/XiaomiMiMo/MiMo-7B-RL-0530}.

\begin{figure}[t]
    \centering
    \includegraphics[width=0.80\textwidth]{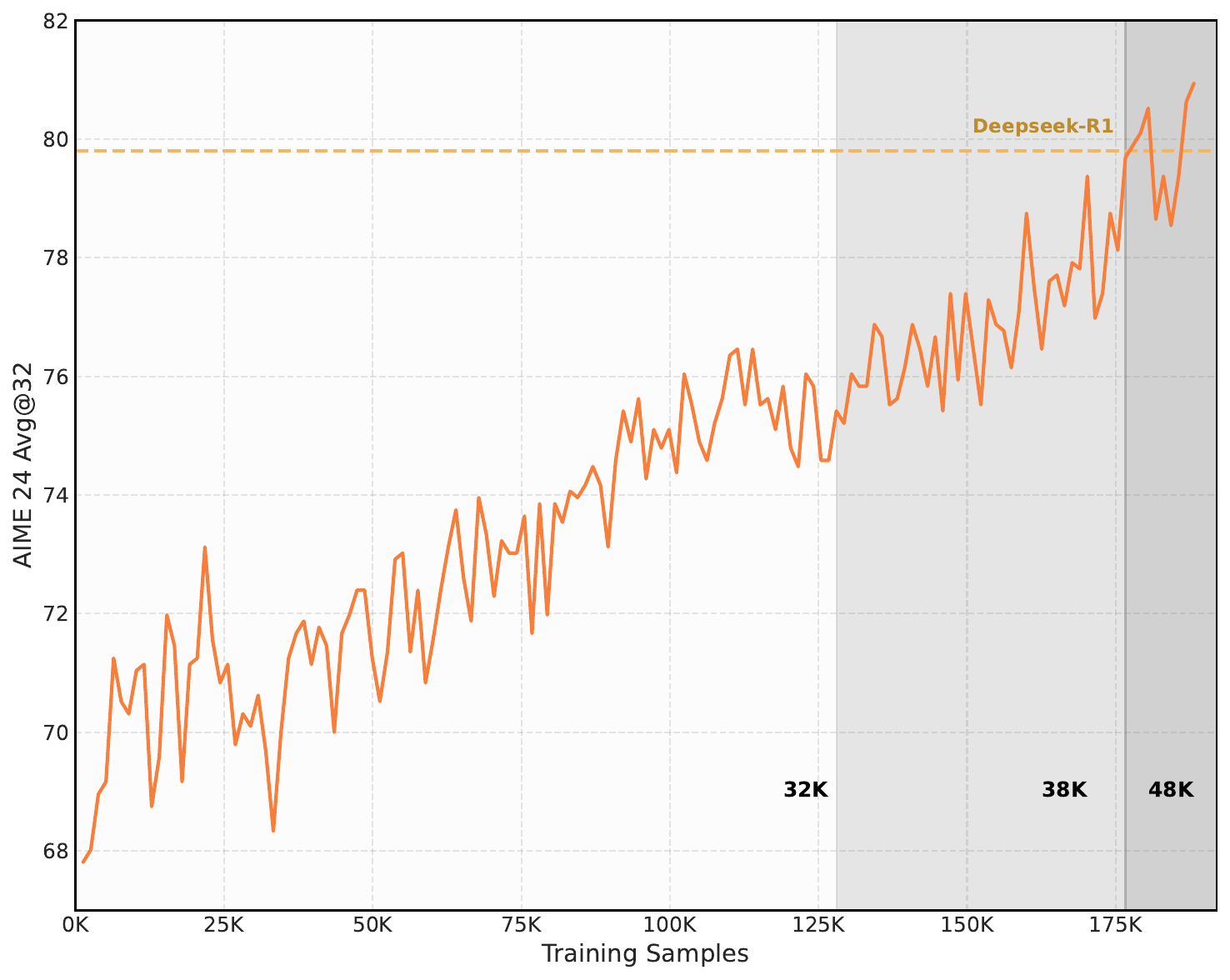}
    \caption{MiMo-7B-RL-0530 Performance Curves on AIME24. }
    \label{fig:mimo_7b_rl_0530_aime24}
\end{figure}

\section{Conclusion}
This work introduces \mimo{}, a series of LLMs which unlock advanced reasoning capabilities through optimized pre-training and post-training process. 
Exposed to diverse reasoning patterns during pre-training, \mimobase{} possesses exceptional reasoning potential, outperforming models of significantly larger scale.
For post-training, with our robust and efficient RL frameworks, we trained \mimozero{} and \mimorl{} which demonstrate superior reasoning capabilities across mathematics, code and general tasks.
We hope this work offers insights for developing more powerful reasoning models.

\bibliography{main}


\appendix

\newpage

\section{Contributions and Acknowledgments}
We would like to express our sincere gratitude to all contributors, including those not listed in the paper, for their invaluable support and efforts. Authors within each role are listed alphabetically by their first name. 

\definecolor{ourblue}{RGB}{0, 0, 0}
\definecolor{ourgreen}{RGB}{0, 0, 0}
\definecolor{ourred}{RGB}{0, 0, 0}

\begin{multicols}{2} %
\noindent
\textbf{\color{ourred} Core Contributors} \\
\color{ourred} Bingquan Xia \\
\color{ourred} Bowen Shen \\
\color{ourred} Cici \\
\color{ourred} Dawei Zhu \\
\color{ourred} Di Zhang \\
\color{ourred} Gang Wang \\
\color{ourred} Hailin Zhang \\
\color{ourred} Huaqiu Liu \\
\color{ourred} Jiebao Xiao \\
\color{ourred} Jinhao Dong \\
\color{ourred} Liang Zhao \\
\color{ourred} Peidian Li \\
\color{ourred} Peng Wang \\
\color{ourred} Shihua Yu \\
\color{ourred} Shimao Chen \\
\color{ourred} Weikun Wang \\
\color{ourred} Wenhan Ma \\
\color{ourred} Xiangwei Deng \\
\color{ourred} Yi Huang \\
\color{ourred} Yifan Song \\
\color{ourred} Zihan Jiang \\

\noindent
\textbf{\color{ourblue} Contributors} \\
\color{ourblue} 
\color{ourblue} Bowen Ye \\
\color{ourblue} Can Cai \\
\color{ourblue} Chenhong He \\
\color{ourblue} Dong Zhang \\
\color{ourblue} Duo Zhang \\
\color{ourblue} Guoan Wang \\
\color{ourblue} Hao Tian \\
\color{ourblue} Haochen Zhao \\
\color{ourblue} Heng Qu \\
\color{ourblue} Hongshen Xu \\
\color{ourblue} Jun Shi \\
\color{ourblue} Kainan Bao \\
\color{ourblue} Kai Fang \\
\color{ourblue} Kang Zhou \\
\color{ourblue} Kangyang Zhou \\
\color{ourblue} Lei Li \\
\color{ourblue} Menghang Zhu \\
\color{ourblue} Nuo Chen \\
\color{ourblue} Qiantong Wang \\
\color{ourblue} Shaohui Liu \\
\color{ourblue} Shicheng Li \\
\color{ourblue} Shuhao Gu \\
\color{ourblue} Shuhuai Ren \\
\color{ourblue} Shuo Liu \\
\color{ourblue} Sirui Deng \\
\color{ourblue} Weiji Zhuang \\
\color{ourblue} Weiwei Lv \\
\color{ourblue} Wenyu Yang \\
\color{ourblue} Xin Zhang \\
\color{ourblue} Xing Yong \\
\color{ourblue} Xing Zhang \\
\color{ourblue} Xingchen Song \\
\color{ourblue} Xinzhe Xu \\
\color{ourblue} Xu Wang \\
\color{ourblue} Yihan Yan \\
\color{ourblue} Yu Tu \\
\color{ourblue} Yuanyuan Tian \\
\color{ourblue} Yudong Wang \\
\color{ourblue} Yue Yu \\
\color{ourblue} Zhenru Lin \\
\color{ourblue} Zhichao Song \\
\color{ourblue} Zihao Yue \\

\end{multicols} %


\end{document}

%% file: tables/base_eval.tex
\begin{table}[htbp]
    \centering
    \small
    \resizebox{0.9\textwidth}{!}{
    \begin{tabular}{l c | c c c | c}
    \toprule
    \multirow{2}{*}{\textbf{Benchmark}} & \multirow{2}{*}{\textbf{\# Shots}} & \textbf{Llama-3.1} & \textbf{Gemma-2} & \textbf{Qwen2.5} & \textbf{\textbf{MiMo-}} \\
    & & \textbf{8B Base} & \textbf{9B Base} & \textbf{7B Base} & \textbf{7B Base} \\
    \midrule
    \multicolumn{6}{l}{\textbf{General}} \\
    BBH {\tiny (EM)} & 3-shot & 64.2 & 69.4 & 70.4 & 75.2 \\
    GPQA-Diamond {\tiny (EM)} & 5-shot & 33.3 & 24.2 & 35.4 & 25.8 \\
    SuperGPQA {\tiny (EM)} & 5-shot & 19.9$^*$ & 22.6$^*$ & 24.6$^*$ & 25.1 \\
    DROP {\tiny (F1)} & 3-shot & 59.5 & 67.9$^*$ & 61.5$^*$ & 69.2 \\
    MMLU {\tiny (EM)} & 5-shot & 65.3 & 71.2 & 74.2 & 71.2 \\
    MMLU-Redux {\tiny (EM)} & 5-shot & 58.4$^*$ & 67.9 & 71.1 & 65.3 \\
    MMLU-Pro {\tiny (EM)} & 5-shot & 37.1 & 44.7 & 45.0 & 41.9 \\
    ARC-Easy {\tiny (EM)} & 25-shot & 84.3 & 88.3 & 86.4 & 85.2 \\
    ARC-Challenge {\tiny (EM)} & 25-shot & 57.7 & 68.2 & 63.8 & 62.3 \\
    HellaSwag {\tiny (EM)} & 10-shot & 82.0 & 81.9 & 80.4 & 80.0 \\
    PIQA {\tiny (EM)} & 0-shot & 80.3 & 81.9 & 78.5 & 79.4 \\
    WinoGrande {\tiny (EM)} & 5-shot & 60.5 & 73.9$^*$ & 75.9 & 78.0 \\
    RACE-High {\tiny (EM)} & 5-shot & 44.3 & 48.3 & 46.8 & 44.1 \\
    TriviaQA {\tiny (EM)} & 5-shot & 70.6 & 76.5 & 60.0 & 60.8 \\
    NaturalQuestions {\tiny (EM)} & 5-shot & 27.7 & 29.2 & 24.1 & 24.5 \\
    AGIEval {\tiny (EM)} & 0-shot & 38.2$^*$ & 21.6$^*$ & 44.4 & 48.3 \\
    \midrule
    \multicolumn{6}{l}{\textbf{Mathematics}} \\
    AIME 2024 {\tiny (Pass@1)} & 0-shot & 0.3$^*$ & 0.0$^*$ & 10.1$^*$ & 32.9  \\
    AIME 2025 {\tiny (Pass@1)} & 0-shot & 0.0$^*$ & 0.0$^*$ & 4.3$^*$ & 24.3  \\
    GSM8K {\tiny (EM)} & 8-shot & 48.5$^*$ & 70.2$^*$ & 80.2$^*$ & 75.2 \\
    MATH {\tiny (EM)} & 4-shot & 16.9$^*$ & 36.4$^*$ & 44.3$^*$ & 37.4 \\
    \midrule
    \multicolumn{6}{l}{\textbf{Code}} \\
    LiveCodeBench v5 {\tiny (Pass@1)} & 0-shot & 0.4$^*$ & 0.0$^*$ & 5.0$^*$ & 32.9 \\
    HumanEval {\tiny (Pass@1)} & 1-shot & 37.8$^*$ & 41.5$^*$ & 56.7$^*$ & 51.8 \\
    HumanEval+ {\tiny (Pass@1)} & 1-shot & 31.7$^*$ & 31.1$^*$ & 50.0$^*$ & 44.5 \\
    MBPP {\tiny (Pass@1)} & 3-shot & 58.4 & 63.9 & 76.7 & 69.2 \\
    MBPP+ {\tiny (Pass@1)} & 3-shot & 49.9 & 52.9 & 64.2 & 56.6 \\
    CRUXEval-I {\tiny (EM)} & 2-shot & 41.5 & 49.8 & 52.4 & 47.6 \\
    CRUXEval-O {\tiny (EM)} & 2-shot & 36.8 & 42.4 & 48.5 & 56.3 \\
    \midrule
    \multicolumn{6}{l}{\textbf{Chinese}} \\
    C-Eval {\tiny (EM)} & 5-shot & 52.2 & 57.0 & 81.8 & 68.7 \\
    CMMLU {\tiny (EM)} & 5-shot & 52.1 & 58.4 & 82.7 & 70.9 \\
    \bottomrule
    \end{tabular}
    }
    \caption{
        Comparison among \mimobase{} and other open-source base models of comparable size.
        Results marked with * are obtained using our internal evaluation framework.
    }
    \label{tab:base_eval}
\end{table}

%% file: sections/rlinfra.tex
\subsection{RL Infrastructures}

We develop the \rlsys{} and enhance vLLM’s robustness to enable efficient dynamic-sampling-based RL training.
We construct our RL system based on verl~\citep{sheng2024hybridflow}, an open-source RL training library.
The library uses Ray~\citep{moritz2018ray} to manage computation and communication, implementing the rollout and training phases in Ray Actors and exchanging training data through Ray Objects.
Although verl supports flexible implementations of various RL algorithms, it suffers from GPU idle time during both rollout and reward computation phases.
Due to the skewness in response lengths, we observe that most GPUs remain idle while waiting for a few long-sequence rollout workers, resulting in wasted computational resources and a slow training process.
Several prior works have identified this issue and proposed system-level solutions~\citep{zhong2024rlhfuse,team2025kimi,seed2025seed}. 
However, most of these solutions rely on asynchronous training, which modifies the underlying algorithm and introduces staleness in long-sequence responses.
Rule-based reward computation is also time-consuming, particularly for code data, leading to idle periods for valuable GPU resources.
Our use of dynamic sampling, while improving sample efficiency, exacerbates GPU idle time, and leads to wasted examples during multi-turn rollouts.
To simultaneously optimize GPU utilization and reduce sample waste, we develop the \rlsys{}, opportunistically filling sample batches into rollout while performing asynchronous reward computation.
Our system builds on the vLLM inference engine~\citep{kwon2023efficient}, and we collaborate with the open-source community to enhance the robustness of vLLM's ``external launch'' mode within the verl framework. 
Additionally, we implement MTP in vLLM to support both \mimo{} and \mimorl{}.

\begin{figure}[t]
    \centering
    \includegraphics[width=0.82\linewidth]{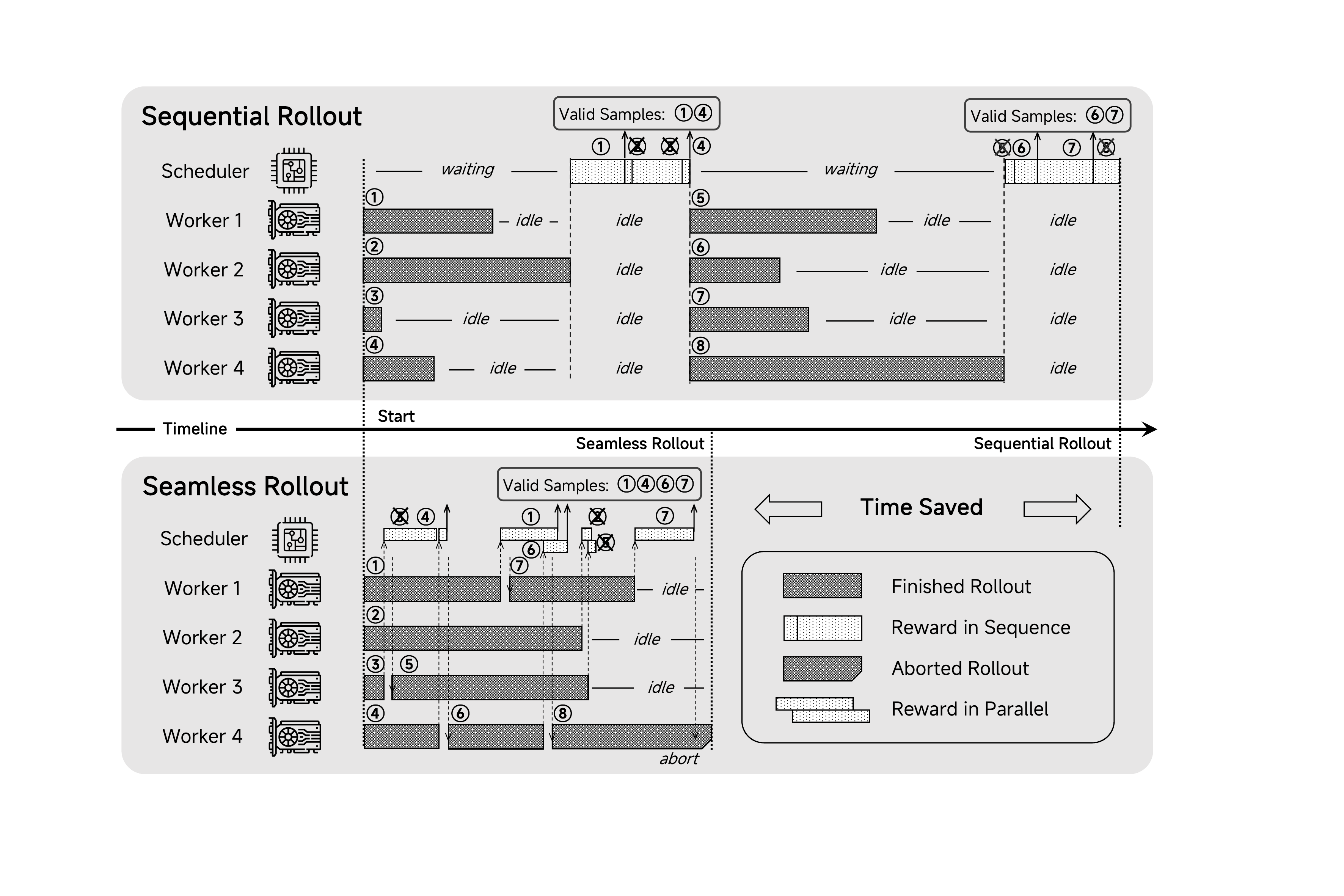}
    \caption{An overview of the \rlsys{} for \mimorl{}.}
    \label{fig:rlinfra}
\end{figure}

\subsubsection{\rlsys{}}

\rlsys{} optimizes GPU utilization in rollout workers through efficient task scheduling, minimizing idle time during continuous operation.
The engine consists of the following components: (a) continuous rollout, (b) asynchronous reward computation, and (c) early termination.
It achieves a $2.29\times$ speedup in training and a $1.96\times$ speedup in validation.

\paragraph{Continuous Rollout}
The core of \rlsys{} lies in proactively handling completed rollout tasks and initiating new rollouts.
Unlike naive dynamic sampling implementations that delay reward computation until all rollout workers complete, \rlsys{} eliminates synchronization barriers between generation and reward phases.
It actively monitors completed workers, immediately computes their rewards, and triggers new rollouts on demand.
After computing rewards, we update the number of valid samples and the current step's pass-rate statistics, then launch new rollout tasks if active tasks are insufficient to meet training demands based on these statistics.
As illustrated in Figure~\ref{fig:rlinfra}, the \rlsys{} initiates a new task upon completing rollout tasks \ding{174}\ding{175}\ding{172}\ding{177} to meet demand, whereas after finishing tasks \ding{173}\ding{176}\ding{178}, it predicts that ongoing tasks are sufficient and thus schedules no additional ones.

\paragraph{Asynchronous Reward Computation}
While reward computation for math data is rapid, judging code-related data incurs significant overhead, leading to prolonged GPU idle time. 
Additionally, the sequential nature of naive reward computation fails to utilize the multiprocessing capabilities of modern processing units. 
To resolve these issues, we employ Ray to launch asynchronous reward computation, which facilitates concurrent management of rollout and reward tasks.
Upon task completion, the system dynamically forwards rollout outputs for reward evaluation or aggregates results to update the sample state, as shown in Figure~\ref{fig:rlinfra}.
Dedicated servers are allocated for code-specific reward computation to prevent bottlenecks in the rollout pipeline. 

\paragraph{Early Termination}
When the number of valid samples exceeds the required training batch size, careful management of ongoing tasks becomes essential. 
Abrupt termination of ongoing tasks tends to suppress the generation of long-sequence responses, which could destabilize RL training dynamics. 
A straightforward solution involves waiting for all active tasks to complete before randomly sampling required batch from the outputs. 
However, this approach may extend waiting times if a long-sequence rollout initiates near the end of the dynamic sampling phase. 
To mitigate this delay while preserving data distribution integrity, we implement a first-in-first-out selection strategy.
We terminate ongoing tasks only if the valid sample count meets the batch requirement and all tasks initiated prior to these selected samples have completed.
In Figure~\ref{fig:rlinfra}, the last rollout is aborted since earlier samples already reach the required batch size.

\input{tables/rlinfra_train}

\paragraph{Experimental Analysis}
We randomly choose a 5-step training trace to evaluate the performance of \rlsys{}.
The experiment is conducted on 256 H20 GPUs, and the results are presented in Table~\ref{tab:rlinfra:train}.
``Overall Speedup'' measures end-to-end RL training efficiency; ``Rollout Speedup'' shows the acceleration of rollout and reward tasks; ``Normalized GPU Idle Time'' reflects the total idle GPU hours.
The above metrics are normalized with respect to the naive dynamic sampling implementation.
``GPU Idle Ratio'' quantifies the average proportion of GPU inactivity during rollout and reward computation; ``Sample Waste Ratio'' represents the ratio of excess valid samples generated relative to the required batch size.
In \rlsys{}, aborted tasks are considered in GPU idle time.

All three components contribute to faster dynamic sampling and smaller GPU idle time.
Though the experiment without dynamic sampling can achieve higher throughput, it incurs significant sample inefficiency due to numerous zero-gradient training samples.
These zero-gradient samples not only diminishes the effective training batch size, but also risk destabilizing the training dynamics of the RL algorithm.
Given an average sample pass rate of 41\% within this 5-step experiment, static sampling achieves a sample efficiency similar to naive dynamic sampling; the latter does not train zero-gradient data but incurs wasted samples.
Equipped with all three components, \rlsys{} achieves a comparable one-step training time compared to static sampling while demonstrating superior sample efficiency.
The sample pass rate of 41\% leads to a sample waste ratio of 22\% in the naive implementation; in practice, this ratio can be larger in different situations.
Through continous rollout and dynamic launch scheduling, \rlsys{} reduces the sample waste ratio to around 15\%.

\paragraph{Accelerated Validation}
During validation, we can directly stream the rollout and reward tasks using \rlsys{}.
Similar to the naive implementation, currently we set the validation batch size equal to the dataset length and launch all rollout tasks simultaneously.
Our implementation utilizes asynchronous reward computation, achieving a 1.96$\times$ speedup while reducing idle GPU time to 25\%, as demonstrated in Table~\ref{tab:rlinfra:val}.
Notably, the experimental results demonstrate \rlsys{}'s potential for static sampling, which also has one-pass rollout and reward computation.
If the validation dataset is sufficiently large, further acceleration can be achieved by optimizing the batch size for validation and employing continuous rollout.

\input{tables/rlinfra_val}

\subsubsection{vLLM-based Inference Engine}

Our RL system employs vLLM~\citep{kwon2023efficient} as the inference engine. 
To accommodate our model's new features, we have extended the framework with additional functionalities.

\paragraph{MTP Support}
As described in Section~\ref{sec:pretrain:modelarc}, our models integrate MTP modules to enhance performance. 
We have implemented and open-sourced MTP support for our models, enabling efficient inference for MTP-equipped architectures.

\paragraph{Better Robustness}
In verl, vLLM is deployed using the \textit{external launch} mode, which may show instability in some scenarios.
We've enhanced engine robustness to address these issues.
We clear computed blocks in \textit{prefix caching} during pre-emption to maintain KVCache consistency.
We disable asynchronous output processing when increasing the number of scheduler steps to ensure compatibility and optimize performance.

%% file: tables/rlinfra_train.tex
\begin{table}[htbp]
    \centering
    \small
    \begin{tabular}{l|ccccc}
    \toprule
    \textbf{Method} & \makecell{\textbf{Overall}\\\textbf{Speedup $\uparrow$}} & \makecell{\textbf{Rollout}\\\textbf{Speedup $\uparrow$}} & 
    \makecell{\textbf{Normalized}\\\textbf{GPU Idle Time $\downarrow$}} & \makecell{\textbf{GPU Idle}\\\textbf{Ratio $\downarrow$}} & \makecell{\textbf{Sample}\\\textbf{Waste Ratio $\downarrow$}} \\
    \midrule
    w/o Dynamic Sampling & 2.45$\times$ & 2.82$\times$ & 0.36 & 70.8\% & / \\
    \midrule
    Naive Dynamic Sampling & 1.00$\times$ & 1.00$\times$ & 1.00 & 69.3\% & 22.1\% \\
    + Continous Rollout & 1.99$\times$ & 2.20$\times$ & 0.25 & 38.8\% & 13.9\% \\
    + Async. Reward & 2.09$\times$ & 2.34$\times$ & 0.21 & 34.0\% & 16.4\% \\
    + Early Termination & 2.29$\times$ & 2.61$\times$ & 0.15 & 27.7\% & 12.9\% \\
    \bottomrule
    \end{tabular}
    \caption{
        The experimental results of \rlsys{} compared with baseline methods. 
    }
    \label{tab:rlinfra:train}
\end{table}

%% file: tables/rlinfra_val.tex
\begin{table}[htbp]
    \centering
    \small
    \begin{tabular}{l|ccc}
    \toprule
    \textbf{Method} & \textbf{Speedup $\uparrow$} & \textbf{Normalized GPU Idle Time $\downarrow$} & \textbf{GPU Idle Ratio $\downarrow$} \\
    \midrule
    Naive Validation & 1$\times$ & 1 & 65.8\% \\
    \rlsys{} & 1.96$\times$ &  0.25 & 32.9\% \\
    \bottomrule
    \end{tabular}
    \caption{
        The validation speedup and GPU idle time of the naive implementation and the \rlsys{}. The experiment is conducted on 256 H20 GPUs using our full validation dataset.
    }
    \label{tab:rlinfra:val}
\end{table}

%% file: tables/rl_eval.tex
\begin{table}[t]
    \centering
    \small
    \resizebox{\textwidth}{!}{
    \begin{tabular}{l | c  c | c  c  c  c |c}
    \toprule
    \multirow{2}{*}{\centering \textbf{Benchmark}}  & \textbf{GPT-4o} & \textbf{Claude-3.5-} & \textbf{OpenAI}& \textbf{QwQ-32B} & \textbf{R1-Distill-} & \textbf{R1-Distill-} & \textbf{MiMo-}\\
    & \textbf{0513} & \textbf{Sonnet-1022} & \textbf{o1-mini} & \textbf{Preview} & \textbf{Qwen-14B} & \textbf{Qwen-7B} &\textbf{7B-RL} \\
    \midrule
    \multicolumn{8}{l}{\textbf{General}} \\
    GPQA Diamond {\tiny (Pass@1)}& 49.9 & 65.0 & 60.0 & 54.5 & 59.1 & 49.1 & 54.4 \\
    SuperGPQA {\tiny (Pass@1)} & 42.4 & 48.2 & 45.2 & 43.6 & 40.6 & 28.9 & 40.5 \\
    DROP {\tiny (3-shot F1)}  & 83.7 & 88.3 & 83.9 & 71.2 & 85.5 & 77.0 &  78.7 \\
    MMLU-Pro {\tiny (EM)} & 72.6 & 78.0 & 80.3 & 52.0 & 68.8 & 53.5 &  58.6 \\
    IF-Eval {\tiny (Prompt Strict)} & 84.3 & 86.5 & 84.8 & 40.4 & 78.3 & 60.5 &  61.0 \\
    \midrule
    \multicolumn{8}{l}{\textbf{Mathematics}} \\
    MATH500 {\tiny (Pass@1)} & 74.6 & 78.3 & 90.0 & 90.6 & 93.9 & 92.8 &  95.8 \\
    AIME 2024 {\tiny (Pass@1)}  & 9.3 & 16.0 & 63.6 & 50.0 & 69.7 & 55.5 & 68.2 \\
    AIME 2025 {\tiny (Pass@1)}  & 11.6 &  7.4 & 50.7 & 32.4 & 48.2 & 38.8 & 55.4 \\
    \midrule
    \multicolumn{8}{l}{\textbf{Code}} \\
    LiveCodeBench v5 {\tiny (Pass@1)} & 32.9 & 38.9 & 53.8 & 41.9 & 53.1 & 37.6 & 57.8 \\
    LiveCodeBench v6 {\tiny (Pass@1)} & 30.9 & 37.2 & 46.8 & 39.1 & 31.9 & 23.9 & 49.3 \\
    \bottomrule
    \end{tabular}
    }
    \caption{
        Comparison between \mimorl{} and other representative models. 
    }
    \label{tab:rl_eval}
\end{table}